\documentclass{article}

\usepackage[preprint]{corl_2024} 
\usepackage[utf8]{inputenc} 
\usepackage[T1]{fontenc}    
\usepackage{hyperref}       
\usepackage{url}            
\usepackage{booktabs}       
\usepackage{amsfonts}       
\usepackage{nicefrac}       
\usepackage{microtype}      
\usepackage{multirow}
\usepackage{amsmath}
\usepackage{amssymb}
\usepackage{bbm}
\usepackage{graphicx}
\usepackage{caption}
\usepackage{subcaption}
\usepackage{wrapfig}
\usepackage{colortbl}
\usepackage{adjustbox}
\usepackage{siunitx}
\usepackage{cite}
\usepackage{textcomp}
\usepackage[inline]{enumitem}
\usepackage{mathtools}
\usepackage[bb=dsserif]{mathalpha}
\usepackage{algorithm}
\usepackage{algpseudocode}

\newcommand{\defeq}{\vcentcolon=}
\newcommand{\Real}{\mathbb{R}}
\newcommand{\SO}{\mathrm{SO}}
\newcommand{\so}{\mathrm{so}}
\newcommand{\SE}{\mathrm{SE}}
\newcommand{\se}{\mathrm{se}}
\newcommand{\R}{\mathrm{R}}
\newcommand{\trans}{\mathrm{t}}
\newcommand{\T}{\mathrm{T}}
\newcommand{\I}{\mathrm{I}}
\newcommand{\N}{\mathcal{N}}

\newcommand{\norm}[1]{\left\lVert #1 \right\rVert}
\newcommand{\card}[1]{\left\lvert #1 \right\rvert}

\newcommand{\mycomment}[1]{}

\title{Towards Explaining Uncertainty Estimates \\ in Point Cloud Registration}

\author{
  Ziyuan Qin$^{1}$, 
  Jongseok Lee$^{2,3}$, 
  Rudolph Triebel$^{2,3}$
  \\
  $^1$ Technical University of Munich (TUM)
  \quad
  $^2$ Karlsruhe Institute of Technology (KIT)
  \\
  $^3$ Institute of Robotics and Mechatronics, German Aerospace Center (DLR)  
  \\
  \texttt{ziyuan.qin@tum.de, \{jongseok.lee, rudolph.triebel\}@dlr.de} \\
}

\begin{document}
\maketitle

\begin{abstract}
Iterative Closest Point (ICP) is a commonly used algorithm to estimate transformation between two point clouds. The key idea of this work is to leverage recent advances in explainable AI for probabilistic ICP methods that provide uncertainty estimates. Concretely, we propose a method that can explain \textit{why} a probabilistic ICP method produced a particular output. Our method is based on kernel SHAP (SHapley Additive exPlanations). With this, we assign an importance value to common sources of uncertainty in ICP such as sensor noise, occlusion, and ambiguous environment. The results of the experiment show that this explanation method can reasonably explain the uncertainty sources, providing a step towards robots that know when and why they failed in a human interpretable manner. 
\end{abstract}

\section{Introduction}

Point cloud registration plays an essential role in many tasks in robotics and computer vision, such as simultaneous localization and mapping~\citep{izadi2011kinectfusion}, grasping~\citep{targo}, and augmented reality~\citep{Wu2018ar}. Iterative Closest Point (ICP) is a widely used algorithm to register two point clouds~\citep{besl1992icp}. Given the source and reference point clouds and an initial transformation estimate, the ICP iteratively minimizes the Euclidean distance between pairs of matching points from both point clouds.

In practice, the ICP pose estimation process is usually affected by several sources of uncertainty. These include sensor noise, initial pose uncertainty, partial overlap, under-constrained situations, and intrinsic ICP randomness~\citep{censi2007accurate, maken2021stein, brossard2020new}. Due to the error incurred by these error sources, a single pose estimate is often insufficient to obtain a robust transformation and accurately localize a moving object. Thus, many ICP algorithms provide not only the point estimate of transformation but also the uncertainty estimate for the transformation parameters~\citep{censi2007accurate, brossard2020new, maken2021stein, maken2020bayesian, landry2019cello}. Depending on how these sources of uncertainty vary, the uncertainty also varies.

Given the uncertainty of the pose estimate produced by any uncertainty-aware point cloud registration algorithm, this work uses the SHAP kernel to build a model-agnostic explanation module and identify the association between uncertainty sources and estimated uncertainty in point cloud registration~\citep{lundberg2017shap}, so that after removing or mitigating the sources, the pose estimates achieve a lower uncertainty and arrive at a better localization. To the best of our knowledge, this work is the first attempt to show that uncertainty estimates in point cloud registration can be explained in a human interpretable manner.

\section{Background on Explanation Methods}
\label{sec:explanation-method}

To attribute uncertainty to various sources, we rely on the existing explanation method. In this section, we provide background on the additive feature attribution method, LIME, Shapley values, and eventually how the kernel SHAP combines these ideas into a linear model with desirable properties.

\textbf{Additive Feature Attribution Method:} Let $f$ be the initial model to be explained and $g$ the explanation model. Local methods explain a prediction $f(x)$ using a single input $x$. These methods use \textit{simplified} inputs $x'$, which are mapped to the \textit{original} inputs $x$ by a function $x = h_x(x')$. The goal is to ensure $g(z') \approx f(h_x(z'))$ whenever $z' \approx x'$ \citep{lundberg2017shap}.
Methods like LIME~\citep{ribeiro2016lime} and SHAP~\citep{lundberg2017shap} follow the additive feature attribution framework, represented by a linear function of binary variables:

\begin{equation}
\label{eq:additive_fa}
g(z') = \phi_0 + \sum_{i = 1}^M \phi_i z_i',
\end{equation}
where $z' \in \{0,1\}^M$, $M$ is the number of simplified features, and $\phi_i \in \mathbb{R}$. 
According to Equation (\ref{eq:additive_fa}), explanation models assign an effect $\phi_i$ to each feature. Summing the feature attributions $\phi_i$ and the bias $\phi_0$ approximates the output $f(x)$ of the original model.

\textbf{LIME (Local Interpretable Model-agnostic Explanations):}  LIME~\citep{ribeiro2016lime} is proposed to satisfy three criteria: interpretable (qualitative understanding between input and response), local fidelity (explanation locally approximates model near prediction) and model-agnostic (explanation method applicable to \textit{any} model). When using a local linear explanation model, LIME follows Equation~(\ref{eq:additive_fa}), and is therefore an additive feature attribution method. To find $\phi$, LIME minimizes the following objective function with respect to $x$:
\begin{equation}
\label{eq:local_lime}
\mathop{{\arg\min}\vphantom{\sim}}\limits_{\displaystyle _{g \in G}}  ~ L(f, g, \pi_{x'}) + \Omega(g),
\end{equation}
where $G$ is a set of interpretable models, $L(f, g, \pi_{x'})$ measures the unfaithfulness of $g$ in approximating $f$ locally, and $\Omega(g)$ captures the complexity of the explanation. By minimizing $L$ weighted by the local kernel $\pi_{x'}$ and controlling the complexity with $\Omega$, LIME ensures both local fidelity and interpretability.

\textbf{Kernel SHAP:} In cooperative game theory, players cooperate in a coalition and receive a certain payoff. Depending on their individual contribution to the total payoff, \textbf{Shapley value} assigns a contribution to each player. This idea in game theory can easily be extended to model explanations, where each player represents a feature, and the coalition value signifies model prediction~\citep{lipovetsky2001analysis, vstrumbelj2014explaining, datta2016algorithmic}.

There are three desirable properties with additive feature attribution methods:
\begin {enumerate*}[label=\arabic*)]
\item local accuracy, \item missingness, and \item consistency.
\end {enumerate*}
Surprisingly, SHAP (SHapley Additive exPlanation) values, proposed by~\citep{lundberg2017shap}, provide the unique additive feature importance measure that adheres to the three properties. While precise calculation of SHAP values can be computationally expensive, the subsequent Kernel SHAP method achieves similar approximation accuracy with fewer evaluations of the original model.

Proposed and proved in~\citep{lundberg2017shap}, \textbf{Shapley kernel} can induce Shapley values consistent with Equation (\ref{eq:additive_fa}) and the three properties to solve Equation (\ref{eq:local_lime}):
\begin{subequations}\label{eq:kernel}
\begin{align}
\Omega(g) &= 0 \label{eq:kernel_complexity} \\
\pi_{x'}(z') &= \frac{(M-1)}{\binom{M}{|z'|} |z'| (M - |z'|)} \label{eq:kernel_weight} \\
L(f,g,\pi_{x'}) &= \sum_{z' \in Z} \left( f(h_x(z')) - g(z') \right)^2 \pi_{x'}(z') \label{eq:kernel_loss}
\end{align}
\end{subequations}
Given that $g(z')$ is assumed to follow a linear form in Equation (\ref{eq:additive_fa}) and $L$ represents a quadratic loss, Equation (\ref{eq:local_lime}) can be solved via \textit{weighted linear regression}. The weighting function $\pi_{x'}$ forms the basis of the Shapley kernel, and the entire process of weighted linear regression is termed the \textbf{kernel SHAP}.

\section{On the Problem of Explainable ICP}

In this section, we first present the background on ICP, the sources of uncertainty in ICP, and existing uncertainty techniques. Then, we present our main goal and motivate why we should explain uncertainty in ICP.

\subsection{Iterative Closest Point}
\label{sec:ICP}
ICP aligns a source point cloud with a target point cloud by estimating the rigid transformation (rotation and translation) between them~\citep{rusinkiewicz2001efficient}. Since the correct point correspondences are unknown and the searching process is not differentiable, it is generally impossible to determine the optimal rotation and translation in one step. Thus, iterations are performed, each consisting of two steps:

\textbf{First step:} Using an initial pose $\theta = \{x, y, z, \text{roll}, \text{pitch}, \text{yaw}\}$, the point correspondences are established by minimizing the distance function. There are two variants of the distance function: point-to-point~\citep{besl1992icp} and point-to-plane~\citep{chen1992object}. The commonly used point-to-point distance metric is:    \begin{equation*}
        \text{point-to-point}(s_i', \mathcal R) = \min_{r_j \in \mathcal R} \left\Vert r_j - s_i' \right\Vert,
        \label{eq:correspondences}
    \end{equation*}
    where $s_i, r_j \in \Real^3$ are the corresponding pair of points in 3D space belonging to source cloud $\mathcal{S} = \{s_i\}_{i=1}^{N}$ and reference cloud $\mathcal{R} = \{r_j\}_{j=1}^{M}$ respectively. $s_i' = (\R s_i + \trans)$ is a transformed point in the source cloud, $\trans \in \Real^{3 \time 1}$ is a translation vector consisting of $\theta_{1: 3}$ and $\R \in \mathbb{R}^{3 \times 3}$, parametrized by $\theta_{4: 6}$, represents a rotation matrix.
\textbf{Second step:} Estimate the relative rigid transformation $\R$ and $\trans$ by minimizing the distance between each pair of corresponding points. The point-to-point cost function has the following form:
    \begin{equation*}
        \operatorname*{argmin}_{\theta}  \mathcal L(\theta) = \frac{1}{N}\sum_i^N  ||(\R \, \mathbf{s_i}+ \trans) - \mathbf{r_j}||^2.
        \label{eq:icp-objective}
    \end{equation*}

The estimated rigid transformation consists of rotation $\R$ and translation $\trans$. $\R$ is in the special orthogonal group $\SO(3) \defeq \left\{\R \in \Real^{3 \times 3} \mid \R^{\top} \R=I, \det(\R) = 1\right\}$ and $\trans \in \Real^3$. Combining both rotation and translation, in homogeneous coordinate, the rigid transformation lives in the group of special Euclidean transformations: $\SE(3) \defeq \left\{
        g = \begin{bmatrix}
            \R & \trans \\
            0 & 1
        \end{bmatrix}
        \mid \R \in \SO(3), \trans \in \Real^3 \right\} \subset \mathbb{R}^{4 \times 4}$.

\subsection{Sources of ICP Uncertainty}
\label{sec:ICP-uncertainty-sources}

According to the literature~\citep{censi2007accurate, maken2021stein, brossard2020new}, the uncertainty of ICP uncertainty comes mainly from five possible sources: \textbf{sensor noise}, \textbf{initial pose uncertainty}, \textbf{partial overlap}, \textbf{under-constrained situations}, \textbf{intrinsic ICP randomness}.

\textbf{Sensor noise}: the sensor noise consists of both sensor white noise and sensor bias noise. Sensor white noise is caused when each point measured in a point cloud is influenced by an independent random sensor noise, a function of the point depth and beam angle~\citep{wang2018characterization}. Sensor bias noise is induced when the observed points share common errors caused by environmental conditions such as the temperature drift effect, distortion due to sensor calibration, and the physical nature or texture of the perceived material~\citep{pomerleau2012noise}.

\textbf{Initial pose uncertainty}: If the initialization of the pose has a large uncertainty, the solution would converge to a local minimum rather than the attraction area of the true solution~\citep{landry2019cello, brossard2020new}.

\textbf{Partial overlap}: partial overlap refers to a situation where there is a limited amount of overlap between the points in the cloud of sources and target points. This limitation may arise from the viewpoint of the sensor, occlusions, or the relative motion between the sensor and the scene. It can be challenging to establish reliable correspondences between the two point clouds, leading to alignment uncertainty~\citep{maken2021stein, chen2020overlapnet}.
    
\textbf{Under-constrained situations}: an ambiguous point cloud structure such as a rotational symmetric bottle would yield a high uncertainty in rotation. This type of uncertainty can cause erroneous data associations and induce a high pose estimation error~\citep{censi2007accurate}.
    
\textbf{Intrinsic ICP randomness}: though the ICP process itself is deterministic, a certain degree of uncertainty is introduced by the random filtering process, e.g., sub-sampling and outlier rejection, so that even two solutions with exactly the same inputs could differ~\citep{pomerleau2015review}.

The last two sources, \textbf{under-constrained situations}\footnote{For instance, when a cylindrical mug removes its handle, it becomes a rotationally symmetric shape, and the resulting uncertainty would inflate.} and \textbf{intrinsic ICP randomness}, however, they are not explained for the sake of simplicity\footnote{\textbf{Intrinsic ICP randomness} could hardly be quantified, whereas \textbf{under-constrained situations} lack real-world observations.}. In this work, we focus on the first three sources of uncertainty: \textbf{sensor noise}, \textbf{initial pose uncertainty}, and \textbf{partial overlap}.

\subsection{Uncertainty Estimation in ICP}
\label{sec:uncertainty_estimation}

There exist many ICP algorithms that either produce a single transformation estimate like the original ICP~\citep{besl1992icp} or are uncertainty-based, which provide a distribution of possible poses and estimate the pose uncertainty~\citep{pomerleau2015review, maken2021stein}. In this work, we utilize ICP to demonstrate that the explanation method can indeed produce reasonable estimates. That is, if our explanation works for the original ICP producing a single pose estimate, it could easily adapt to any other ICP algorithm, single estimate or not. Multiple pose estimates are sampled to compute the uncertainty estimate using the Kullback-Leibler (KL) divergence. More details are discuss in our experiments later.

\subsection{Problem Statement}

We have looked into how ICP functions, what may cause them to fail, and uncertainty quantification techniques. Our goal is to not only estimate the uncertainty but also explain the uncertainty. We argue that in ICP, the explanations should be in human interpretable concepts, which are the sources of ICP uncertainty in our scenario. Several advantages can exist. First, knowing the causes of uncertainty, we can perform active perception at a symbolic level, e.g., at a robotic task level, we might be able to use robot controls to eliminate ICP failures. Furthermore, such information can be used in a teleoperation setup. A human operator might be able to react to perception failures better if the operator is informed about what is causing the failures.



\section{On How to Perturb the Three Uncertainty Sources}

Sensor noise and partial overlap are perturbed by altering the input point clouds, whereas initial pose uncertainty is perturbed by transforming the initial poses.

\subsubsection{Sensor Noise}
\label{sec:sensor_noise}
As described in Section \ref{sec:ICP-uncertainty-sources}, sensor noise is comprised of sensor white noise and sensor bias noise. Since the sensor white noise is more straightforward to model, it is represented as a zero-mean Gaussian noise without loss of generality. The noise is then added to each point in the two input point clouds.

\subsubsection{Initial Pose Uncertainty}
\label{sec:intial_pose}

As derived in~\citep{brossard2020new}, given a small rotation $\alpha, \beta, \gamma$ around the $x$, $y$, $z$ axes respectively, the full rotation $\R$ can be linearly approximated as:
\begin{equation*}
    \R \approx
    \begin{bmatrix}
    1 & -\gamma & \beta \\
    \gamma & 1 & -\alpha \\
    -\beta & \alpha & 1
    \end{bmatrix}
    = \I_3 + 
    \begin{bmatrix}
        \alpha \\
        \beta \\
        \gamma
    \end{bmatrix}
    = \I_3 + \delta^\wedge
\end{equation*}
where the operator $\wedge$ turns the vector $\delta = \begin{bmatrix}
\alpha & \beta & \gamma
\end{bmatrix}^T$ in $\Real^3$ to a $3 \times 3$ skew-symmetric matrix in Lie algebra $\so(3)$, the tangent space to Lie group $\SO(3)$.

Combined with translation $\rho$, the rigid transformation can be linearized as $\T \approx \I_4 + \xi^\wedge$ with
\begin{equation*}
    \xi^\wedge
    \defeq \begin{bmatrix}    
        \delta^\wedge & \rho \\
        0 & 0
    \end{bmatrix} \in \mathbb{R}^{4 \times 4}, \,
    \xi=\begin{bmatrix}
        \delta \\
        \rho
    \end{bmatrix}
\end{equation*}
where $\delta \in \Real^3$ represents rotation, $\rho \in \Real^3$ signifies translation, and operator $\wedge$ turns $\xi \in \Real^6$ to a $4 \times 4$ real matrix in Lie algebra $\se(3)$, the tangent space to Lie group $\SE(3)$.

If $\xi$ is taken randomly as a Gaussian $\xi \sim \N(0, \Sigma)$, where $\Sigma \in \Real^{6 \times 6}$ is the covariance matrix, then $I_4 + \xi^\wedge$ defines a small transformation. For the ground truth pose $\T_{gt}$, the initial guess is a randomly perturbed pose close to $\T_{gt}$: $\T_{init} = \T_{gt} (\I_4 + \xi^\wedge) = \T_{gt} + \T_{gt} \xi^\wedge$. Using the notion of concentrated Gaussian distribution~\citep{bourmaud2015continuous},
\begin{equation}
\label{eq:concentrated_gaussian}
    \T_{init} = \T_{gt} \exp(\xi^\wedge)
\end{equation}
where $\xi \sim \N(0, \Sigma) \in \Real^6$, and the $\exp(\cdot)$ operator denotes the exponential map from $\se(3)$ to $\SE(3)$.

Note that $\xi$ can be viewed as the error between $\T_{gt}$ and $\T_{init}$. Indeed, the relative $\SE(3)$ transformation between the ground truth and initial poses is encoded in $\xi$ as $\exp(\xi^\wedge) = \T_{gt}^{-1} \T_{init}$.

\subsubsection{Partial Overlap}
\label{sec:partial_overlap}

For a pair of input point clouds $P_1^l$ and $P_2^l$, an overlapping ratio can be computed similar to that in~\citep{chen2020overlapnet}. Given ground truth absolute pose $\T_1$ and $\T_2$, the point clouds are transformed from local camera frame to world frame: $P_1$ and $P_2$. For each point $p$ in $P_2$, the nearest neighbor $NN(p)$ in $P_1$ is found using kNN with $k = 1$. The neighbor is valid only if the Euclidean distance between $p$ and $NN(p)$ is no larger than $d = 0.2 \, \text{m}$. Calculating the ratio between the number of valid neighbors and number of points in $P_1$ yields the overlap ratio:
\begin{equation*}
    O_{1, 2} = \frac{N}{\card{P_1}} \text{, where } N = \sum_{p \in P_2} \mathbb{1}(\norm{NN(p) - p} \leq d).
\end{equation*}

In reality, the means to perturb input point clouds $P_1$ and $P_2$ with partial overlap are various and depend on the specific scenario, a simple but effective way to reduce overlapping region is as follows. Given the current overlap ratio $O_{1, 2}$, $P_2$ is perturbed to reach target overlap ratio $O'_{1, 2} = O_{1, 2} - \lambda$, with $\lambda > 0$. By simple calculation, $N - \card{P_1} \cdot O'_{1, 2}$ points need to be removed from the overlapping region in $P_2$ to reduce current overlap ratio by $\lambda$.

\section{The Proposed Method}
\label{sec:method}

We note that sensor noise and partial overlap affect the source and reference point clouds, whereas initial pose uncertainty influences the range of sampled initial poses. As mentioned in Section \ref{sec:ICP}, ICP takes the input point clouds and initial poses and produces a set of pose estimates. Finally, uncertainty is estimated using KL divergence in Section \ref{sec:uncertainty_estimation}. Given different levels of perturbation, the input point clouds and initial poses would change, and the ICP algorithm would yield different uncertainty estimates. Kernel SHAP is then used to explain how important each uncertainty source contributes to the uncertainty estimate. As kernel SHAP uses the additive feature attribution method in Equation (\ref{eq:additive_fa}), the explanation is:
\begin{equation}
    g(z') = \phi_{sn} z_{sn}' + \phi_{ip} z_{ip}' + \phi_{po} z_{po}'
\end{equation}
where $z_{sn}', z_{ip}', z_{po}' \in \{0,1\}$ represent simplified input features, $\phi_{sn}, \phi_{ip}, \phi_{po} \in \mathbb{R}$ refer to the attributed importance of initial pose uncertainty, sensor noise, and partial overlap, respectively. For an instance $i$ and uncertainty source $j$, $z_{ij}' = 1$ signifies that the feature is present, and input is perturbed in this fashion, whereas $0$ indicates the absence of uncertainty source $j$, thus not perturbed by such feature and is thus the reference value. Note that the bias $\phi_0$ is omitted because the expected value for the unperturbed case is $0$, i.e., all $z_{ij}' = 0$ for all $j \in \{1, \ldots, M\}$.

\begin{figure}
    \centering
    \includegraphics[width=\textwidth]{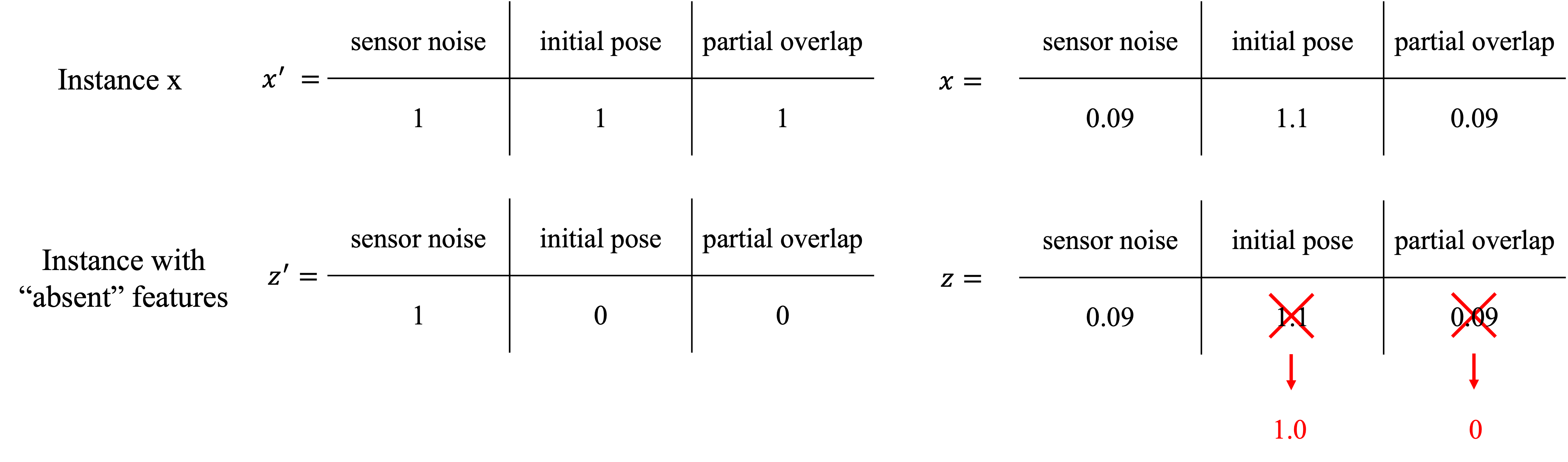}
    \caption{The simplified input on the left, represented by binary conditions, are mapped to the feature values on the right via the mapping function $h_x$.}
    \label{fig:mapping}
\end{figure}

In Algorithm \ref{alg:kernel_shap}, given a perturbation setting $x = \{x_{sn}, x_{ip}, x_{po}\}$ and $M = 3$, for each $i$ in $2^M = 8$, $z'[i]$ is mapped to $z[i]$ via $h_x$, as shown in Figure~\ref{fig:mapping}. Then, uncertainty $y[i]$ is predicted by the ICP algorithm and weight $w[i]$ are calculated by Equation (\ref{eq:kernel_weight}). Finally, the quadratic loss $L$ in Equation~(\ref{eq:kernel_loss}) is minimized by the optimal SHAP values $\phi$ calculated via the closed form solution for weighted linear regression.

\begin{algorithm}
\caption{Kernel SHAP estimates feature contribution for one prediction}\label{alg:kernel_shap}
\begin{algorithmic}
\Require $f$: ICP algorithm that estimates uncertainty, $x = \{x_{sn}, x_{ip}, x_{po}\}$: instance to explain, $M$: number of features is $3$, $r$: reference values are $\{0, 1, 0\}$ (see details in Section \ref{sec:uncertainty_sources_range})
\For{each coalition $z'[i, j] \in \{0, 1\}$, $i\in\{1, \ldots, 2^M\}$, $j \in \{1, \ldots, M\}$}
    \State $z[i, j] \gets h_x(z'[i, j])$ \Comment{$z'[i, j] = 1$ maps $z[i, j]$ to $x[j]$, and $0$ maps $z[i, j]$ to $r[j]$}
    \State $y[i] \gets f(z[i, :])$ \Comment{Calculate estimated uncertainty with ICP algorithm}
    \State $w[i] \gets \pi_{x'}(z'[i, :])$ \Comment{Calculate weights by Equation~(\ref{eq:kernel_weight})}
\EndFor
\State $\phi = (z'^T \text{diag}(w) z')^{-1} z'^T \text{diag}(w) y$ \Comment{closed-form solution for weighted linear regression}
\State \Return $\phi$
\end{algorithmic}
\end{algorithm}
\section{Experiments}


This section shows the experimental setup and how kernel SHAP explains ICP uncertainty in two experiments. The initial experimental setup includes how much to perturb the three uncertainty sources and how the pose uncertainty is estimated.

\textbf{Ranges of Perturbations for Uncertainty Sources:} To analyze the effect of each uncertainty source: sensor noise, initial pose uncertainty, and partial overlap, a sensible range for each source is specified. 

As the mean bias value for Hokuyo sensor is found as \SI{5}{cm} in~\citep{pomerleau2012noise}, the Gaussian \textit{sensor noise} $\N(0, \sigma)$ is added to the two input point clouds with $\sigma$ from \SI{0}{m} to \SI{0.1}{m}, with a step size of \SI{0.01}{m}. 

As shown in Equation (\ref{eq:concentrated_gaussian}), $\T_{init}$ is perturbed by the small \textit{initial pose uncertainty} $\xi \sim \N(0, s \Sigma)$, which is applied with a scale $s$ from $1$ to $2$, with a step size of $0.1$.

Since the number of points in each point cloud is on the order of $10^5$, removing points in the \textit{partial overlap} region of two point clouds would result in a big change in uncertainty estimate. Thus, $\lambda$, difference between original overlap ratio $O_{1, 2}$ and target overlap ratio $O'_{1, 2}$, is set relatively small. Given $O_{1, 2} > 0.1$, $\lambda$ ranges from $0$ to $0.1$, with a step size of $0.01$.

\textbf{Pose Uncertainty Estimation:} In this work, we estimate the pose uncertainty using the KL divergence between the pseudo-true distribution and the perturbed distribution. The pseudo-true distribution is computed by sampling $100$ ICP pose estimates of \textit{unperturbed} inputs: initial pose sampled with zero-mean Gaussian with $s \Sigma$, where $s = 1$, around the ground truth pose, without sensor noise, and unaltered overlap ratio. In contrast, the perturbed distribution is calculated by sampling $100$ ICP pose estimates of \textit{perturbed} inputs. By comparing the covariance matrices of these two normal distributions via KL divergence, we estimate the uncertainty as a scalar, representing the combined effect of all three uncertainty sources.

\subsection{Different Perturbations for the Same Pair of Input Point Clouds in \textit{Apartment}}
\label{sec:diff_pert_apartment}

In this section, we examine various perturbation sets to understand how kernel SHAP elucidates the sources of uncertainty in point clouds $6$ and $7$ from the \textit{Apartment} sequence in the \textit{Challenging datasets}~\citep{challenging}. In the following, summary plot, waterfall plot, and feature dependence plots are illustrated.

\begin{figure}
    \centering
    \begin{minipage}[b]{0.5\textwidth}
        \centering
        \includegraphics[width=\linewidth]{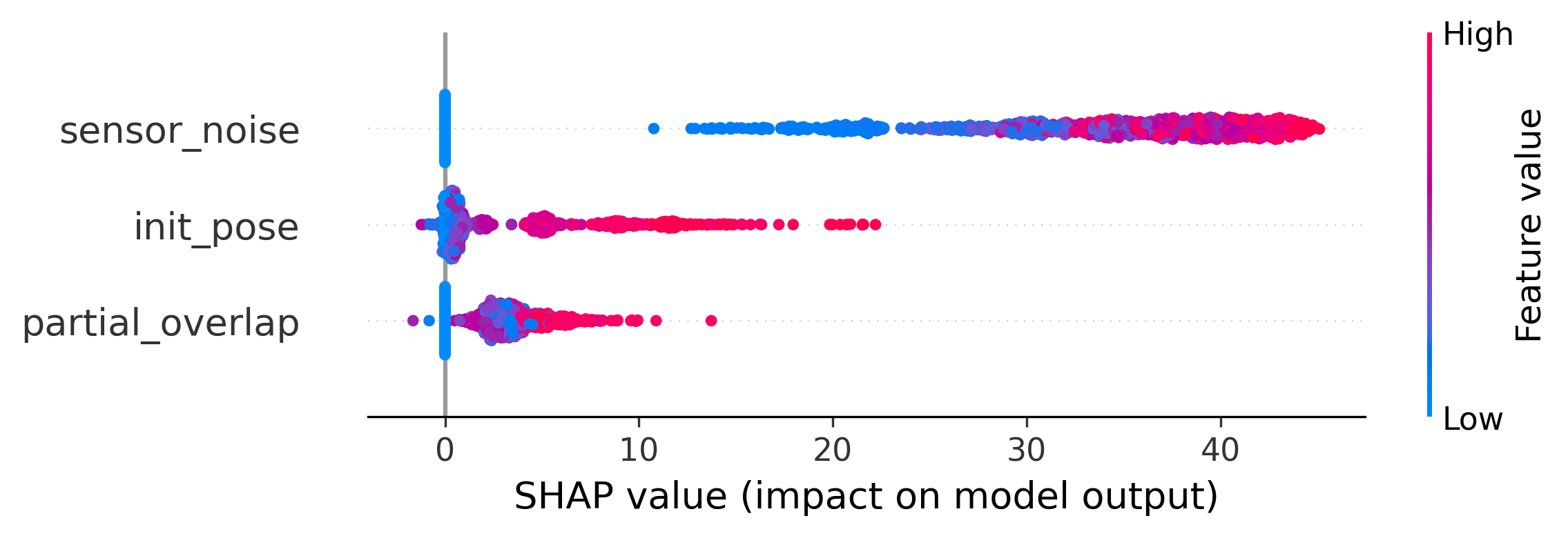}
        \caption{SHAP summary plot.}
        \label{fig:summary_6_7}
    \end{minipage}
    \hfill
    \begin{minipage}[b]{0.46\textwidth}
        \centering
        \includegraphics[width=\linewidth]{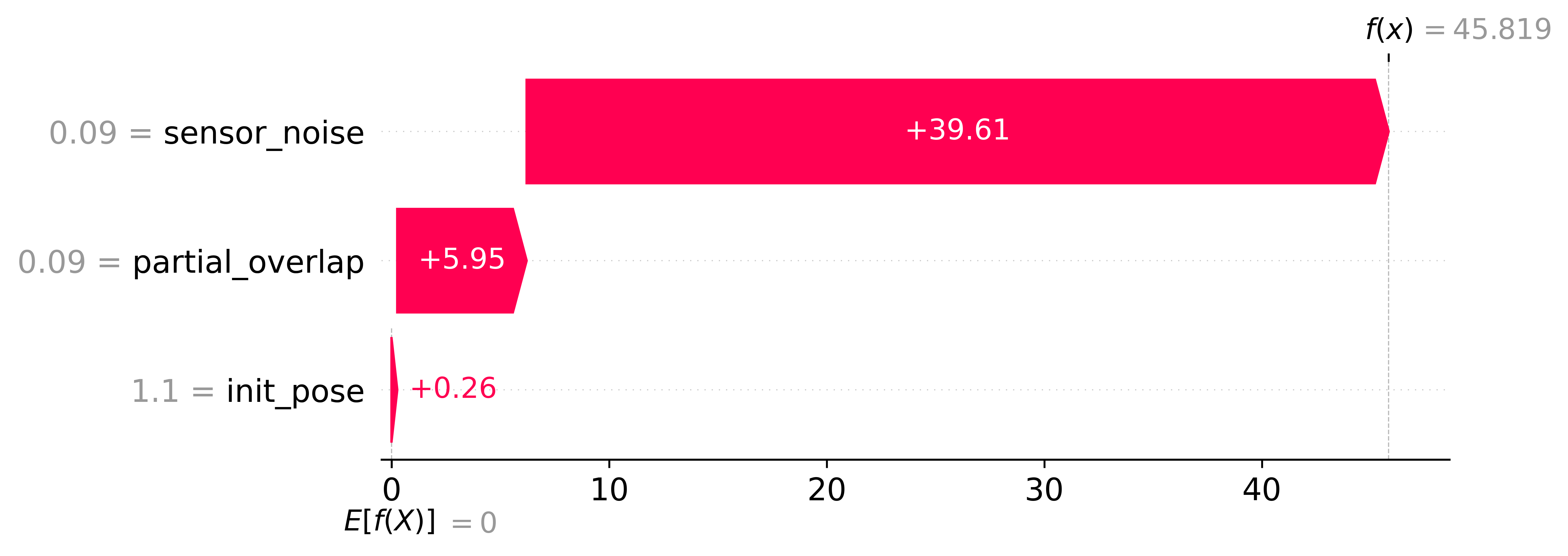}
        \caption{SHAP waterfall plot.}
        \label{fig:waterfall_normal}
    \end{minipage}
\end{figure}

\begin{figure}
    \centering
    \begin{subfigure}[b]{0.32\textwidth}
        \includegraphics[width=\linewidth]{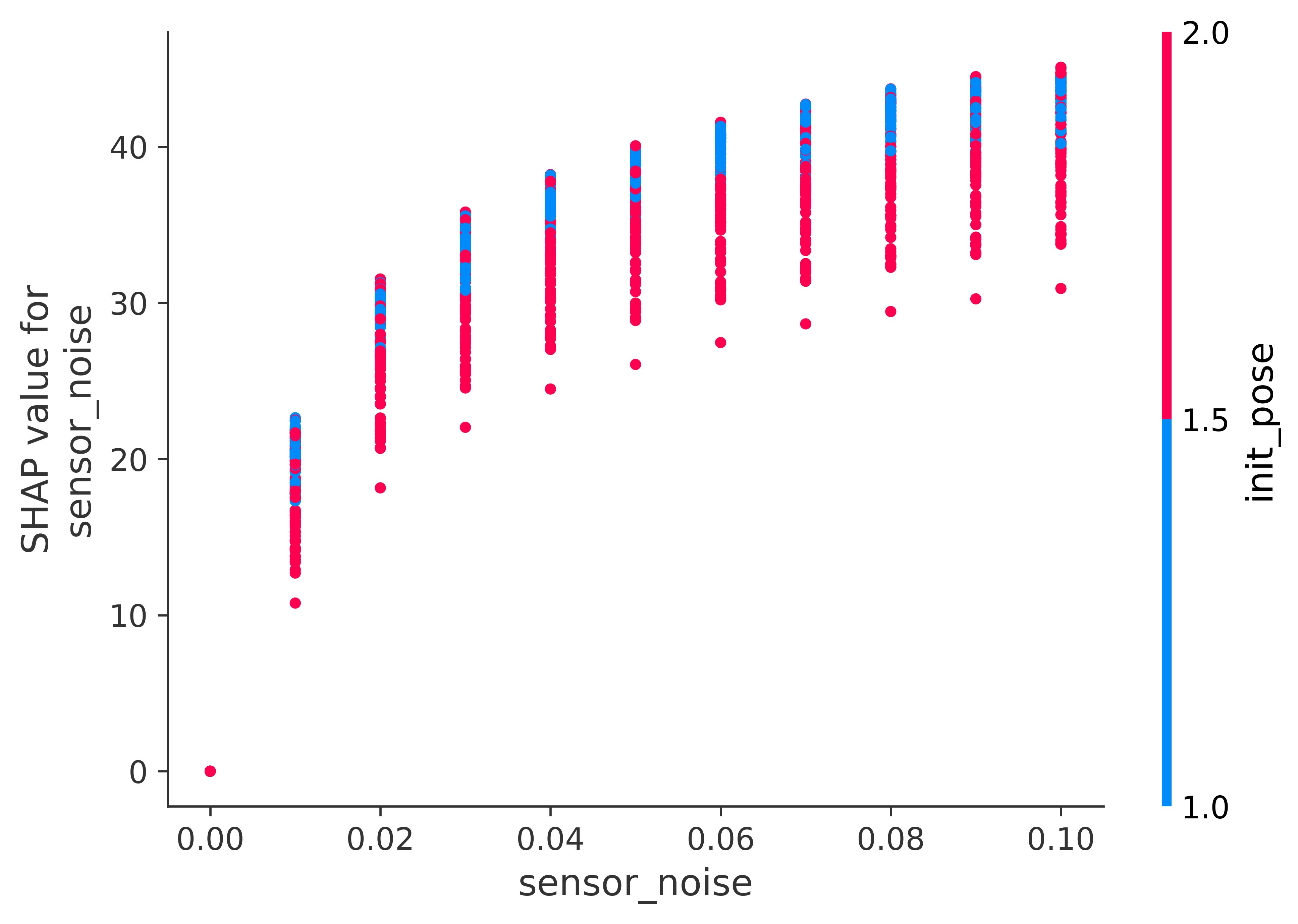}
        \caption{\scriptsize SHAP values of sensor noise vs. feature values of initial pose uncertainty.}
        \label{fig:sn_vs_ip}
    \end{subfigure}
    \hfill
    \begin{subfigure}[b]{0.32\textwidth}
        \includegraphics[width=\linewidth]{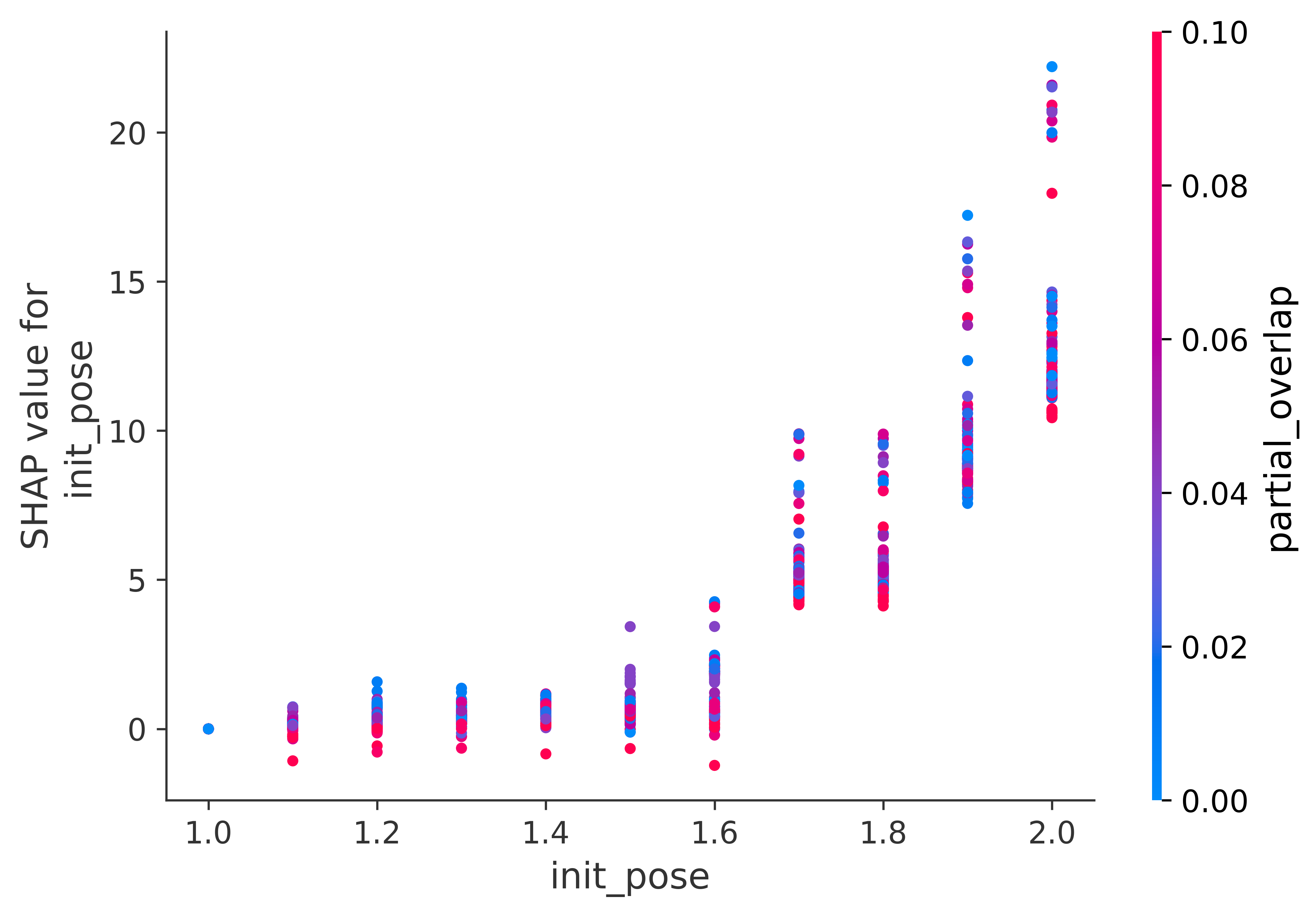}
        \caption{\scriptsize SHAP values of initial pose uncertainty vs. feature values of partial overlap.}
        \label{fig:ip_vs_po}
    \end{subfigure}
    \hfill
    \begin{subfigure}[b]{0.32\textwidth}
        \includegraphics[width=\linewidth]{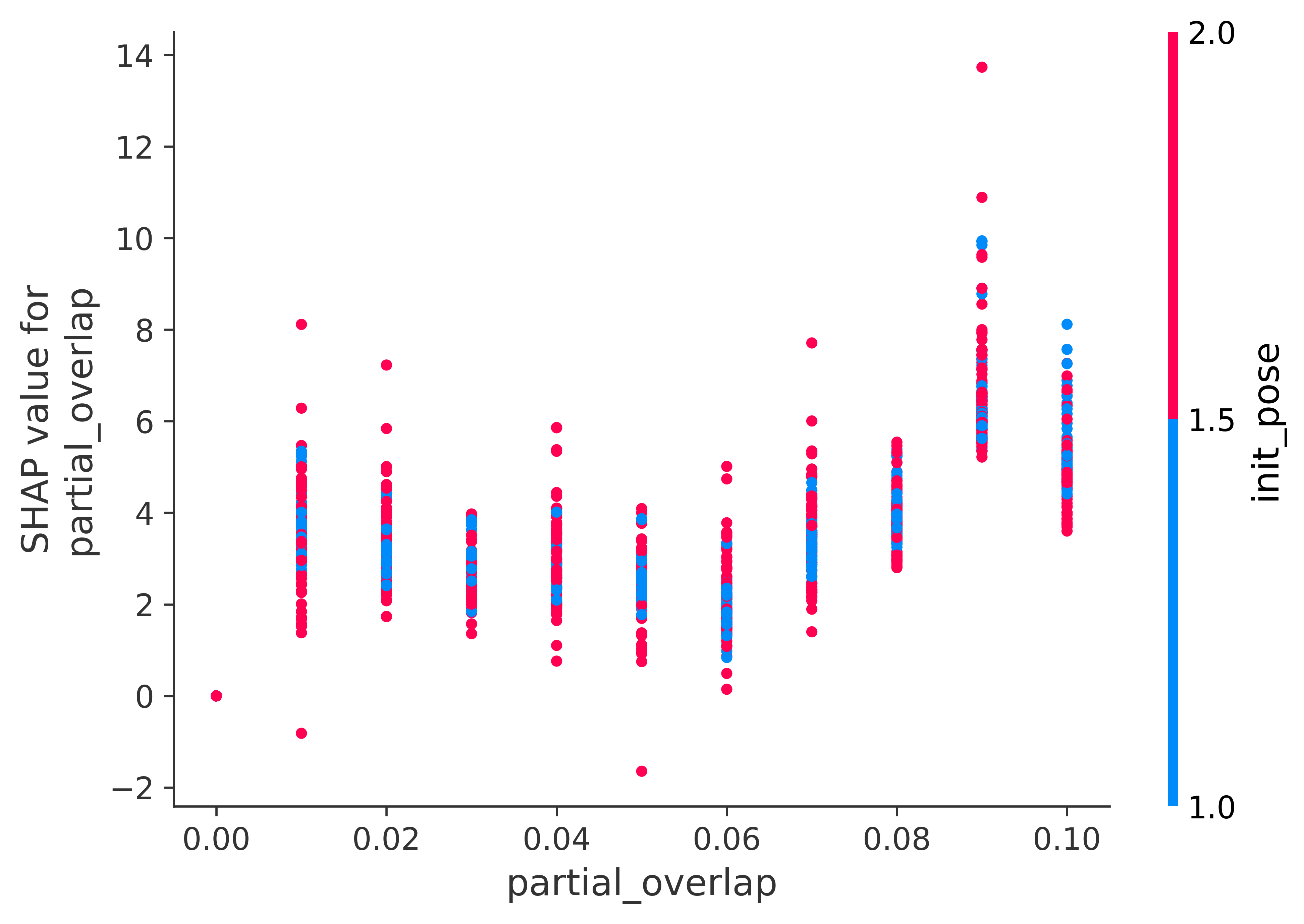}
        \caption{\scriptsize SHAP values of partial overlap vs. feature values of initial pose uncertainty}
        \label{fig:po_vs_ip}
    \end{subfigure}
    \caption{SHAP dependence plots.}
    \label{fig:dependence_6_7}
\end{figure}

\paragraph{Summary plot.} We utilize summary plot to demonstrate the overall impact of each source of uncertainty. In Figure \ref{fig:summary_6_7}, sensor noise correspond to the largest SHAP values, indicating that they contribute the most to pose uncertainty. Positive SHAP values correspond to perturbed features (high feature values, red dots), while when features are unperturbed (low feature values, blue dots), the attributed SHAP value would be close to zero. However, there are a few cases where the SHAP values are negative, as opposed to our predictions.

\paragraph{Waterfall plot.} Waterfall plot shows the contribution of each uncertainty source to individual data instances. In Figure \ref{fig:waterfall_normal}, sensor noise is \SI{9}{cm}, initial pose uncertainty has scale $1.1$, and partial overlap eliminates $9\%$ points in overlapping region of reference point cloud $7$. Among the three uncertainty sources, sensor noise is more important than partial overlap, and initial pose uncertainty has the least influence on uncertainty.

\paragraph{Feature dependence plots.} Feature dependence plots show the relationship between different sources of uncertainty and their corresponding SHAP values. In all three plots in Figure~\ref{fig:dependence_6_7}, an increasing trend of SHAP values vs. feature values could be observed. In Figure~\ref{fig:sn_vs_ip}, for any given sensor noise, high initial pose uncertainty (red dots) seems to reduce the influence of sensor noise, as shown by lower SHAP values. This suggests that when initial pose uncertainty is the dominant factor, it overshadows the influence of sensor noise. In Figure~\ref{fig:ip_vs_po} and Figure~\ref{fig:po_vs_ip}, this interaction effect is less stark.

\subsection{Same Perturbation for Contiguous Pairs of Point Clouds in Different Sequences}
\label{sec:same_pert_all_seq}

In this section, the perturbation of the three uncertainty sources is fixed to the mean value: $\{0.05, 1.5, 0.05\}$. Feature effects are analyzed across contiguous point clouds pairs in all $8$ sequences of \textit{Challenging datasets}. This experimental setup mirrors practical scenarios, where perturbations are fixed and the importance of each uncertainty sources is determined.

After removing outliers using interquartile range, the SHAP values of the three uncertainty sources are analyzed. Table~\ref{tab:median_shapval} displays the median SHAP values for each uncertainty source across all sequences. Except for the sequence $Mountain$, the SHAP values consistently follow the order: \textbf{sensor noise} $>$ \textbf{partial overlap} $>$ \textbf{initial pose uncertainty}. This suggests that, in this perturbation setting, reducing sensor noise would significantly decrease uncertainty across all sequences.

\begin{table}
    \centering
    \begin{tabular}{lccc}
    \hline
    \textbf{Sequence} & \textbf{Sensor Noise} & \textbf{Initial Pose Uncertainty} & \textbf{Partial Overlap} \\
    \hline
    Apartment & 35.571175 & 1.415812 & 2.478877 \\
    ETH & 15.392886 & 0.278460 & 1.279239 \\
    Stairs & 11.755226 & 0.287785 & 0.492705 \\
    Mountain & 3.199889 & \textbf{1.055795} & \textbf{0.973968} \\
    Gazebo Summer & 12.524350 & 0.380334 & 4.750572 \\
    Gazebo Winter & 16.923138 & 2.461525 & 3.904520 \\
    Wood Summer & 11.811761 & 1.138083 & 3.617080 \\
    Wood Autumn & 12.268561 & 1.345611 & 2.197536 \\
    \hline
    \end{tabular}
    \caption{The median SHAP values of the uncertainty sources in all sequences indicate that, except for the sequence \textit{Mountain}, \textbf{partial overlap} is consistently more important than \textbf{initial pose uncertainty} in every other sequence.}
    \label{tab:median_shapval}
\end{table}

While explanation methods assess feature-prediction correlations, causality between uncertainty sources and pose uncertainty estimates remains unsettled. Exploring the causal relationship requires causal inference and is left for future research. In addition, we plan to improve and merge the proposed method to deal with uncertainty in learning-based systems \citep{lee2022trust, lee2020estimating, schnaus2023learning}. Using such methods, we further envision developing a semi-autonomous robotic system \citep{lee2020visual, lee2023virtual} where the human operator can be informed of why the system has failed and take actions to reduce the uncertainty.
\section{Conclusion}

This work employs kernel SHAP to explain the relationship between uncertainty sources and uncertainty estimates. In Section~\ref{sec:diff_pert_apartment}, the effects of each uncertainty source at different perturbation levels are analyzed for the same input point clouds, yielding mostly nonnegative and interpretable SHAP values. In Section~\ref{sec:same_pert_all_seq}, the effects of uncertainty sources under the same perturbation level are studied across contiguous point cloud pairs in all sequences, providing insights into the importance of each source. As kernel SHAP is model-agnostic, it can also explain other ICP algorithms, such as Stein ICP~\citep{maken2021stein}. Kernel SHAP remains applicable even if the range or step of uncertainty sources changes or additional sources, like under-constrained scenarios or ICP randomness, are introduced.




\bibliography{main}  

\end{document}